\documentclass[letterpaper]{article}
\usepackage{aaai}
\usepackage{times}
\usepackage{helvet}
\usepackage{courier}
\usepackage{graphicx}
\usepackage[numbers]{natbib}
\begin{document}
%
\title{Shared Control in Human Robot Teaming: Toward Context-Aware Communication}
\author{Sachiko Matsumoto and Laurel D. Riek\\
Computer Science and Engineering, UC San Diego\\
email: smatsumo@eng.ucsd.edu, lriek@eng.ucsd.edu
}
\maketitle
















\section{Introduction}

In the field of Human-Robot Interaction (HRI) many researchers study shared control systems.
This involves a person and robot ``interacting congruently in a perception-action cycle to perform a dynamic task that either the human or the robot could execute individually under ideal circumstances'' \cite{Abbink2018}.
For example, an agent might assist someone using a robotic arm to grasp an object by identifying the object and moving the arm into a good position to grasp it.
An agent might also assist someone driving a teleoperated robot by preventing them from running into walls or other obstacles.

One of the most important things in shared control is the nature of the communication between the person and robot.
This communication is bi-directional – the agent needs to understand the person (to infer their goals and intentions), and the human needs to understand what decisions the agent is making and why, and how to alter them if/when needed.
Here, interaction is often continuous, where people and agents need to communicate across the duration of a task.

Well-designed communication could improve an operator's situational awareness and increase robot transparency, which can improve performance and help users maintain accurate expectations about the robot's abilities \cite{NAS2021}.
Recent work in explainable robots can help inform transparent communication \cite{Anjomshoae2019}.
Researchers might also draw from the explainable AI field \cite{NAS2021}, though there may be differences when using a physically embodied agent \cite{Setchi2020,Anjomshoae2019}.
Communication also influences understandability (of the robot and task), performance, acceptance of the robot, and so on.

Thus, appropriate, well-designed communication could help the human-robot team handle a variety of challenging situations.
For instance, if the robot was not trained for the task the person wants to use it in, it might be helpful for the robot to communicate its level of uncertainty about the task to the person or how likely it estimates it is to succeed at the task.
This way the person could potentially make more informed decisions about how much to rely on the robot's assistance.

In this paper, we explore key challenges in shared control, in which better communication design could be useful, including when encountering novel situations and contexts, resolving tensions between preferences and performance, and alleviating cognitive burden and interruptions.



\section{Example Shared Control Scenarios}

\begin{figure}
\includegraphics[width=0.5\textwidth]{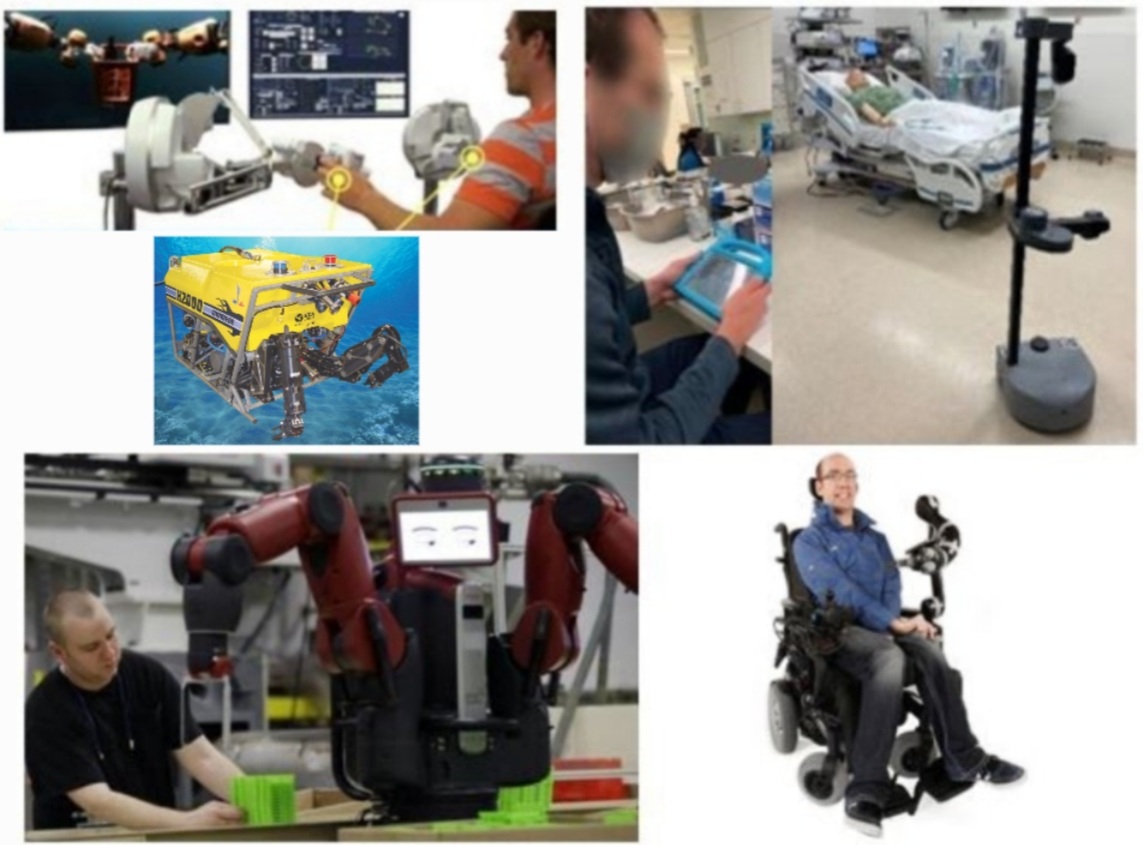}
\caption{We present four different shared control scenarios to explore key challenges in shared control: teleoperating a UUV, teleoperating a telemedicine robot, performing collaborative assembly in a factory, and using an assistive wheelchair mounted robotic arm.}
\label{fig:scenarios}
\end{figure}

Shared control systems can take a variety of forms across several different dimensions.
For instance, they may differ depending on the distance between the person and robot, whether there are other people around the robot or interacting with it, and the criticality and interruptability of the task being performed.
We present some example shared control systems to ground our subsequent discussion of shared control (Fig. \ref{fig:scenarios}).

\textbf{Scenario 1: UUV} A marine biologist might use an unmanned underwater vehicle (UUV) to inspect coral.
In this scenario, the operator and robot are physically distant.
The person cannot physically interact with the robot or the environment the robot is in (except through the robot), and they likely cannot directly see the robot.
Additionally, there are likely no people near the robot.
They might communicate with the agent through a visual (e.g., a GUI) or haptic (e.g., a haptic joystick) interface.

\textbf{Scenario 2: Telemedicine} A healthcare worker might use a mobile telemanipulator robot to examine a patient.
In this scenario, the operator and robot are again physically distant.
However, in this case, the robot is physically proximate to another person and the healthcare worker is expected to interact with them through the robot.
The task also likely is critical and has low interruptability, since interruptions could lead to worse patient outcomes \cite{WernerHolden2015}.

\textbf{Scenario 3: Assembly} A factory worker might use a Baxter robot to collaboratively assemble furniture.
In this scenario, the operator and robot are close to each other, but physically separate.
Thus, unlike in the previous two scenarios, the operator and robot can directly interact with the same environment and directly sense each other (see, hear, etc.).
There are also likely no other people near the operator and robot.

\textbf{Scenario 4: Assistance} A person with a disability might use a wheelchair-mounted arm robot to assist them with tasks such as eating.
As in Scenario 3, the operator and robot are co-located.
However, in this scenario, they are also physically connected.
The operator might use the robot while also interacting with other people.

\section{Key Challenges}

There are several challenges in shared control that researchers should consider.
These challenges may differ between situations, but will likely be of concern in many different use cases.
In particular, we discuss challenges with novel situations, context-aware communication, tensions between performance and preference, and cognitive burden and interruptions.

\textbf{Challenge 1: Novel Situations} The robot may not always be trained for all situations a person wants to use it in.
This is a common problem in robotics, as robots may often be deployed in \textit{open set} conditions where they will encounter new things outside of their training data \cite{Meyer2019,Gutoski2021}.
For example, a person with a disability may want to use the robot to dust a shelf, but the robot might only be trained to assist with picking up objects.
In this case, what should the robot do?
Should it attempt to help still or should it do nothing and allow the person to fully teleoperate the robot?
If the robot attempts to help, it might be able to assist the person, but it could make the task more difficult instead.

The line between what the robot has been trained for and is capable of and what it has not been trained for may not always be clear.
If a telemedicine robot was trained to read patient data off of certain types of patient monitors, and the hospital switches to a new type of patient monitor not in that training set, the robot may or may not still be able to appropriately read the data.

In such scenarios, whether or not the robot should try to assist may depend on the estimated level of risk if an error occurs and the risk tolerance of the person operating it.
In the situation with the telemedicine robot, it is likely the robot should not try to assist with the new monitor, since an error could lead to patient harm.
On the other hand, if the assistive robot tried to help dust the shelf, it likely would not hurt anyone.
It might knock over objects on the shelf, so its assistance might be contingent on how delicate the objects are and the person's comfort level with the possibility of the objects being knocked over.

Robot proficiency and risk self-assessment in such scenarios are still open research areas in robotics.
It is difficult for both robots  and people  to understand what they do not know.
For instance, a robot (and person) may have difficulty estimating the robot's performance on a task it has not tried before.
Thus, the robot's risk calculation may not be very accurate, particularly in complex situations like telemedicine, where the level of risk can change drastically given  environmental factors (e.g., the patient's underlying health conditions, their current vitals, etc.).
In such cases, it may be better for the person to assess the level of risk themselves, rather than rely on an estimate from the robot.

However, people may also produce inaccurate risk assessments, especially if they do not have a good mental model of the robot and a clear understanding of when failures are likely.
For example, say a factory worker uses Baxter to lift a component of the product they are assembling while the worker attaches something underneath it.
If a new model of the product is developed that slightly changes the part where Baxter grips, the person might reasonably assume that Baxter will still be able to reliably grip and lift it, since most people would be able to adapt to the new component.
However, the robot may not be capable of properly gripping it anymore, possibly because of failures of its model for gripping, limitations of its gripper, etc.
The person and robot may need to clearly communicate their proficiencies and risk assessments so they can both gain a more complete picture of the situation and hopefully more accurately assess the risk and whether or not the robot should attempt to help.

Therefore, the robot must somehow communicate to the person that it may not be capable of properly assisting with the task and indicate the level of risk in a clear, easily understandable way.
Some research has indicated that communicating such uncertainty can improve task performance and trust \cite{Chen2021,Barnes2021}.
Researchers should be careful in the design of such communication, as most people do not have an in-depth understanding of statistics, and many different biases can affect the interpretation of risk assessments \cite{Fischhoff1993}.
Additionally, they should be careful to ensure the robot's transparency helps users properly calibrate their trust in the robot \cite{Chen2021,Mercado2016,Stowers2020}, rather than leading them to overtrust the robot or become complacent \cite{Wagner2021,Bhaskara2021} or overwhelmed \cite{Wright2017}.

\begin{figure}
\includegraphics[width=.5\textwidth]{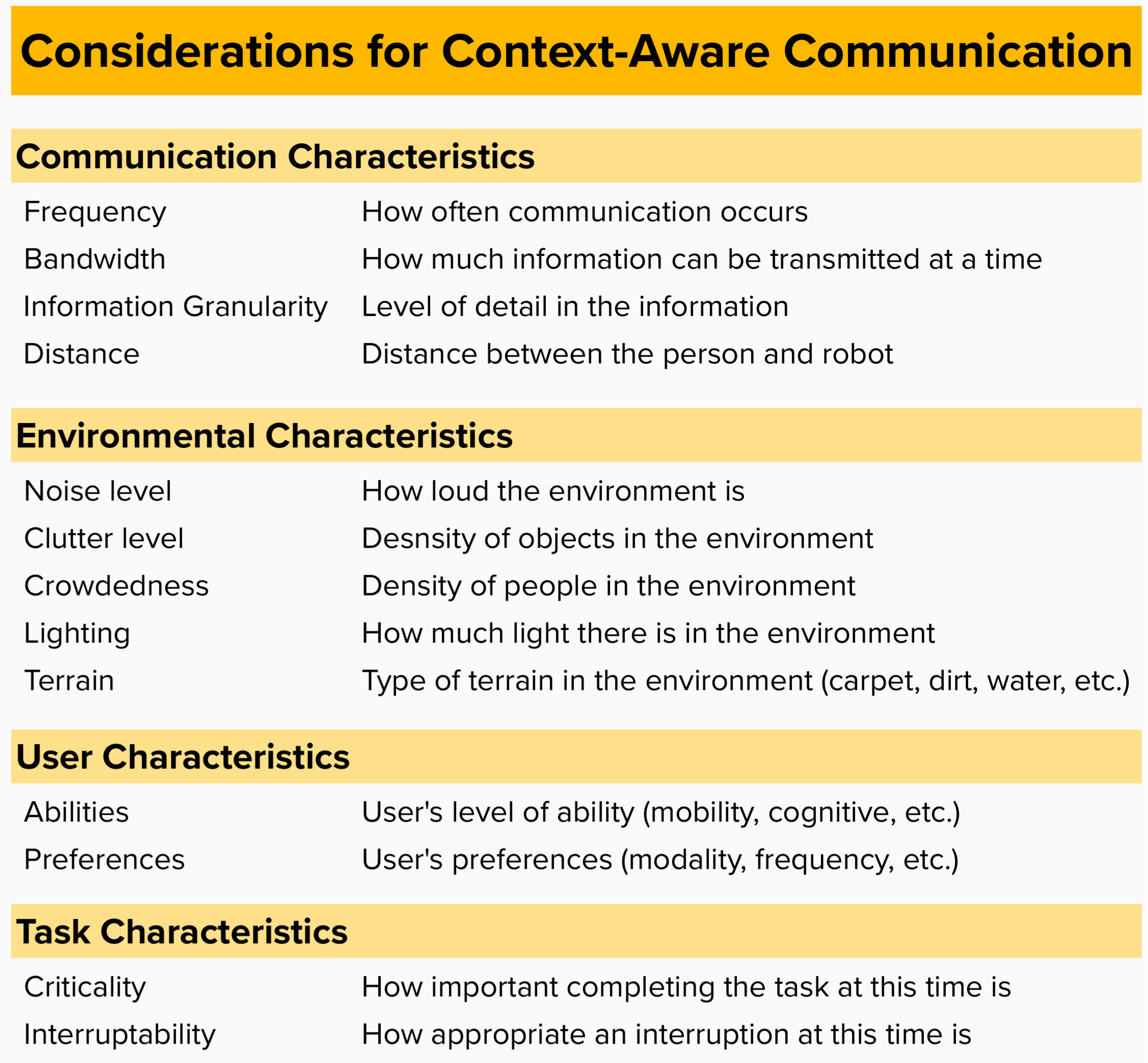}
\caption{Context-aware communication requires considering many different factors, including communication, environmental, user, and task characteristics.}
\label{fig:commFactors}
\end{figure}

\textbf{Challenge 2: Context-Aware Communication} More work is needed to determine how to communicate in a way that is sensitive to the task being performed, the environment it is being performed in, the person it is being performed with, and the ways communication can occur (Fig. \ref{fig:commFactors}).
These factors are interrelated.
For instance, while a marine biologist and healthcare worker might have similar control needs on a telemanipulator robot, their communication and information needs will be quite different. 
For one thing, the marine biologist may not be interacting with other people, so communicating with the robot through audio signals could be appropriate.
In contrast, the healthcare worker might need to talk to a patient, and audio signals from the robot could disrupt that interaction.
Additionally, while both tasks might be critical, the healthcare worker's task may have more serious consequences if they are interrupted during it (e.g., could lead to patient harm).
Shared control systems should consider the criticality and interruptability of a task as part of their communication strategy.

The task context will also affect the type of information people need.
A healthcare worker using a telemedicine robot to conduct  a patient exam may require higher fidelity haptic feedback than a person using an assistive robot to eat food.
While both tasks require some feedback for completion, the healthcare worker might need fine-grained feedback to precisely determine where a patient's pain is coming from or to detect any irregularities in their abdomen.
On the other hand, a person using an assistive robot might just need coarse feedback to determine when the fork at the robot's end effector touches a piece of food and when it pierces it.

The robot's (and person's) task and environmental contexts will also affect how appropriate a given communication method is.
People have limits to the amount of information they can process, and the number of communication channels they can attend to, particularly if they are engaging in a safety-critical task or are under a high cognitive load.
It is important robots in shared control scenarios are well-designed to be able to adapt to these varying conditions to best meet the processing capabilities of the user.

Furthermore, communication will depend on a person's preferences and abilities, which may also depend on the task and environmental context.
For example, if a person is blind or low vision, haptic or aural interaction modalities may be more appropriate than visual modalities.
Our prior work suggests the importance of building systems that can be easily adapted to different contexts depending on the abilities of the user and/or constraints of the environment \cite{Matsumoto2021,Gonzales2015}.
Furthermore, people might prefer different degrees of information granularity in the feedback they receive.
One factory worker working with a Baxter might want detailed information about exactly what step of the assembly process the Baxter is on, while another person might just want to know when the Baxter finishes a set of tasks.

Thus, researchers need to design communication strategies that are able to adapt to the task, environmental and user contexts they are used in.
Such systems should be able to smoothly transition modes as the task or environment changes.

\textbf{Challenge 3: Performance vs. Preference} In shared control systems, the approach that offers the best performance may not always align with the person's preferences.
It is not clear to what extent the robot should prioritize performance over preference, if at all.
Furthermore, how should the robot communicate this trade-off to the person?

The exact relationship between people's preferences, different levels of autonomy, and performance is still unclear, and more research is needed in this area.
In some shared control user studies, participants expressed a preference for a more autonomous robot that performed the task better over a less autonomous robot that performed the task less well \cite{Werner2020,Javdani2018}.
However, in other studies, participants did not prefer more autonomous modes over less autonomous or teleoperated modes, even when they acknowledged more autonomous robot performed better \cite{Kim2012}, took less time or effort to use \cite{Bhattacharjee2020,Javdani2018}, or was safer \cite{Ghorbel2018}.
Some work suggests this preference is influenced by the task difficulty \cite{Dragan2013Policy,Javdani2018}, but other work does not necessarily support this.
For instance, Bhattacharjee et al. \cite{Bhattacharjee2020} and Javdani et al. \cite{Javdani2018} both conducted studies with a feeding scenario, and participants in Bhattacharjee et al. preferred a less autonomous agent mode, while participants in Javdani et al. preferred a more autonomous mode.

If a person's preference does not align with the policy the robot believes will produce the best performance, it is not always clear if the robot should defer to the person's preference.
In many cases, it likely should.
If the task is not critical and errors will not lead to undesirable outcomes, it may not matter if the robot follows the person's preference.
For instance, there are negligible consequences when a marine biologist navigates the robot on a slightly longer path to a patch of coral.

However, in some cases, it is not as clear that the robot should defer to human preference.
A person might prefer less assistance while using a robot to wash themselves.
However, they might not get as clean as they would if the robot provided more assistance, which could lead to adverse health effects.
In these situations, how should the robot behave?

The robot might override the person's preference to increase their safety.
This presents many ethical considerations \cite{Kubota2021}, including whether it is ethical for the robot to override their preferences at all, and raises questions of technological paternalism.
It could also lead to lower adoption of the robot if people are frustrated it does not do what they want.

The robot could also comply with the person's preference.
However, this could give rise to some safety risks or reduced performance for some tasks, particularly when people are cognitively overloaded.

Alternatively, the robot could attempt to communicate to the user the trade-off between their preferences and performance or safety.
The user may not always be aware that their preference lowers performance or safety, so proper communication could keep the user informed about their options and let them know what is happening if the robot did override their preference.
To do this, the robot would need to alert the user that their preferred method is less safe or results in lower performance.
It may also need to explain why this is the case \cite{Setchi2020}.

The robot might also be able to influence the user's actions without explicitly discussing it with the user.
For example, it could attempt to perform the task with the method that increases safety or performance, but defer to the person's preference if the person insists \cite{Nikolaidis2017}.
While it is not guaranteed this will convince the user to change their preferences, it could nudge them towards a method that is safer or performs better.

More research is needed to determine in what situations human preferences are likely to be contrary to the best performance or lower risk.
Additionally, researchers should consider when (if ever) and to what extent robots should ignore people's preferences.
If they do, it will likely be highly dependent on the task and environmental context, as well as the culture of the person using the robot.
In such scenarios, researchers will also need to develop communication methods for the robot to alert the person to the situation.

\textbf{Challenge 4: Interruptions and Cognitive Burden} It is not clear how robots can assist users during interruptions in a task to reduce cognitive burden.
In many environments, people are interrupted frequently.
For instance, in emergency departments, clinicians are interrupted as often as once every four minutes \cite{Weigle2017}.
This adds to the person's cognitive burden and increases the chance of errors.
If robots are introduced into these environments, they will need to be designed with these interruptions in mind.

In some interruption-prone environments, people have tried a variety of methods to reduce interruptions.
For instance, in hospitals, people tried designating ``interruption-free'' zones or using ``do not interrupt'' jackets \cite{Anthony2010,Berdot2021}.
However, it is unclear whether such methods will be feasible or effective long term \cite{Blocker2018}.
With the introduction of robots, new methods will need to be designed to ensure the person operating the robot is not interrupted excessively and that the robot does not disturb other people while they work.

Researchers have explored people's interactions with robots when they are interrupted or multi-tasking.
Often, these studies use interruptions to increase cognitive burden or explore another trait, such as trust \cite{Yang2017}.
More research is needed that specifically focuses on behaviors robots can engage in to mitigate cognitive burden from interruptions and explore strategies robots can use to help users reorient to their original task.

\section{Conclusion}

In this paper, we discussed several key challenges in shared control.
In particular, there are open questions around how the robot should behave in novel situations, ensuring communication is context-aware, balancing preference and performance, and reducing cognitive burden in interruption-prone environments.

In addressing these challenges, researchers should consider ways to improve communication between the person and robot.
While such communication will not resolve all the challenges, it could help keep the person in the loop and make better decisions, regardless of robot behavior.
We hope this paper will draw attention to these challenges and assist researchers in advancing the development of shared control systems.

\section{Acknowledgements}
Work in this paper was supported by the UC MRPI program and the National Science Foundation under Grant No. DRL-2024953.

\bibliographystyle{aaai}
\bibliography{references}

\end{document}